\documentclass[conference]{IEEEtran}

\IEEEoverridecommandlockouts
\usepackage{cite}
\usepackage{amsmath,amssymb,amsfonts}
\usepackage{algorithmic}
\usepackage{graphicx}
\usepackage{textcomp}
\usepackage{xcolor}
\usepackage{hyperref}
\def\BibTeX{{\rm B\kern-.05em{\sc i\kern-.025em b}\kern-.08em
    T\kern-.1667em\lower.7ex\hbox{E}\kern-.125emX}}
\begin{document}

\title{Automatic News Summerization\\
\vspace{10pt}
{\footnotesize \textsuperscript{}\textbf{}

} 

}

\author{%
\textbf{Kavach Dheer}\hspace{0cm}(22kavachdheer@gmail.com) \hspace{0.2cm}
\textbf{Arpit Dhankhar}\hspace{0cm} (arpitdhankhar10@outlook.com) \hspace{0.2cm}

}

\maketitle

\begin{abstract}
Natural Language Processing is booming with its applications
in the real world, one of which is Text Summarization for large texts including news articles.
This research paper provides an extensive comparative evaluation of extractive and abstractive approaches for news text summarization, with an emphasis on the ROUGE score analysis. The study employs the CNN-Daily Mail dataset, which consists of news articles and human-generated reference summaries. The evaluation employs ROUGE scores to assess the efficacy and quality of generated summaries. After Evaluation, we integrate the best-performing models on a web application to assess their real-world capabilities and user experience.
\end{abstract}
\vspace{0.5cm}


\section{Introduction}
In today's rapidly evolving and interconnected world, it is impossible to overstate the importance of news. It plays a crucial function as a catalyst for the development of a society that makes informed decisions. The news is an essential component of democracies because it enables citizens to participate actively in civic affairs. However, the vast quantity of information and the overwhelming nature of news articles can be daunting, posing a challenge for those who want to keep themselves updated.
The application of Natural Language Processing's text summarization [1] models to news articles arises as a potential solution. This method involves condensing extensive news articles into concise formats, thereby presenting the main points in an effective manner. By summarising news articles, individuals are able to remain updated of global events, saving them time and encouraging diverse news consumption. 
Using the ROGUE metric, this study concentrates on extractive and abstraction [1][2] methods of text summarization. The CNN/Daily Mail news dataset will be used to build and train all models.
The research incorporates the most effective news summarization models into a web application that can provide real-time news updates and summarises those. Taking into account the processing and execution time of the models, this study evaluates its ability to produce relevant summaries. The combination of ROGUE evaluation and application testing provides valuable insight into the effectiveness of news summarization models.

Furthermore, we integrated the best-performing news summarization model into a web application capable of fetching live news updates. Through this application, we will evaluate how the selected model performs in real-world scenarios. We will assess its ability to produce accurate and relevant summaries while also considering the processing time required to generate these summaries.

The combination of evaluation through ROGUE scores and real-world application testing will provide valuable insights into the efficiency and effectiveness of news summarization models. By the end of this research, we hope to contribute not only to the advancement of text summarization techniques but also to the practical implementation of a web-based tool that can help individuals consume news, ultimately fostering a more informed and engaged global society.

\section{Litrature Review}
The first paper Mridha et al. [3] gives a thorough review of automatic text summarization that covers many different topics. The paper goes into detail about different methods of text summarization, extractive, abstractive, and hybrid methods. The paper discusses the strengths, flaws, and real-world applications of all of the different methods, as well as how they are evaluated using benchmark datasets which included DUC -2002 and CNN-Daily Mail datasets.

Goularte [4], The authors propose an extractive text summarization approach that incorporates fuzzy rules, catering to automated assessment needs. By employing fuzzy logic, the authors introduce a novel technique that offers increased flexibility and considers more intricate details while generating summaries. The method's applicability in automated assessment systems, which require concise yet accurate summaries of textual content, is a key highlight of their study. The authors outline the implementation of the method and evaluate its performance using DUC- 2002 dataset. The results of the fuzzy method presented the best averages of precision (0.417 and 0.366), recall (0.398 and 0.496), and f-measure (0.406 and 0.421).

Cheng [5] propose a unique text summarization approach that amalgamates the strengths of both extractive and abstractive methods. Their method employs neural networks to identify the most significant sentences and words from the source text, which are then utilized to construct the final summary. By blending extractive and abstractive techniques, their method aims to produce concise, lucid, and highly valuable summaries surpassing the capabilities of either method in isolation.
The researchers introduce a novel neural summarization technique that combines sentence and word extraction techniques. This innovative approach addresses the limitations of conventional summarization methods and offers a promising way to enhance the quality and utility of text summarization through the utilization of neural networks. Additionally, ROUGE evaluation on the DUC- 2002 and 500 DailyMail samples were 21.2,8.3 and 12 for Rogue-1, Rogue-2, and Rogue-L respectively.

Yao [6] introduces a dual encoding framework for abstractive summarization called DEATS((Dual Encoding for Abstractive Text Summarization), incorporating two distinct neural networks: an encoder-decoder network and a pointer-generator network. The encoder-decoder network comprehends the text and generates an initial summary, while the pointer-generator network enables the model to copy relevant phrases from the source text directly, enhancing summary accuracy and information coverage.DEATS achieved a higher Rouge-1 score of 29.91 on the out-of-domain DUC 2004 dataset and a superior Rouge-1 score of 40.85 on the in-domain CNN/DailyMail dataset.

Elbarougy [7], The author introduces a modified PageRank algorithm designed to address the unique characteristics of the Arabic language. The paper emphasizes the importance of text summarization in managing the extensive amount of Arabic textual data, considering challenges such as morphological complexities and linguistic nuances. To overcome these challenges, the researchers propose their modified PageRank algorithm, building upon the original algorithm by Larry Page and Sergey Brin. The modified algorithm incorporates language-specific features, enhancing the ranking process and leading to improved summarization performance for Arabic text.
An evaluation of the proposed approach using  the Essex Arabic Summaries Corpus (EASC) gives the precision, recall, and F- measure were 68.75, 72.92, and 67.99 respectively.

Vaswani [8] proposes the Transformer, a novel architecture for sequence transduction tasks. The Transformer is a recurrent neural network (RNN) model based on attention mechanisms, eschewing convolutions too. Experiments on two machine translation tasks demonstrate that the Transformer achieves state-of-the-art results while being more parallelizable and requiring considerably less training time. The Transformer can learn long-range dependencies between input and output sequences due to the attention mechanism. It has been demonstrated that the Transformer is effective for a variety of sequence transduction tasks, including machine translation, text summarization, and speech recognition.
This paper introduced the attention mechanism which then in the later future is used to create state-of-the-art NLP Models including Generative Pre-trained Transformer (GPT). Though the paper focuses on Machine Translation it made the path for future models to do Text Summerization too .

Lewis [9] introduces BART (Bidirectional and AutoRegressive Transformer), a denoise sequence-to-sequence pre-training method aimed at enhancing language generation, translation, and comprehension tasks. 
The core concept of BART revolves around using denoising autoencoders, a form of unsupervised learning, to train the model before its application. By deliberately perturbing input sequences and training the model to reconstruct the original texts, BART gains precise language representations and discerns crucial language patterns. This denoising pre-training approach equips BART to effectively handle noise and real-world data complexities. The model also employs bidirectional training, combining masked language modeling and left-to-right autoencoding. This versatility enables BART to excel in both auto-regressive and bidirectional tasks. The authors also acknowledge some limitations of the model, such as challenges in dealing with uncommon words and optimization during training. Additionally, the rogue score on CNN/Daily Mail was 44.16, 21.28, and 40.9 on Rogue-1, Rogue-2, and Rogue-L respectively.

Brown's [10] paper  "Language Models are Zero Shot Learners" examines the capabilities of large-scale language models, primarily focusing on the GPT-3 (Generative Pretrained Transformer 3) model. The authors demonstrate that these language models can perform a variety of tasks without task-specific fine-tuning, thereby rendering them zero-shot learners. GPT-3's ability to generate responses to queries, even for tasks it was not explicitly trained on, is surprisingly robust given its large number of parameters and pre-training on diverse datasets.
The paper provides empirical evidence demonstrating that GPT-3 can perform a variety of tasks, including text completion, summarization, question answering, arithmetic, and language translation, among others. In addition, it is capable of "few-shot learning," in which it is provided with only a few examples of a task and then produces correct responses to similar queries. The authors also address the model's limitations, such as sensitivity to phrasing and difficulty with particular query types.

Zhang [11] introduces a new advancement in pre-training techniques for abstractive summarization through the model,  "Pegasus," which implements an innovative approach by utilizing extracted gap sentences to enhance the pre-training process.
The authors found that conventional pre-training methods often face limitations in acquiring the features of abstractive summarization due to limitations in comprehending contextual information. To overcome this, Pegasus adopts a unique strategy of incorporating extracted gap sentences during pre-training, resulting in enhanced summarization capabilities.
Through rigorous evaluation, Pegasus demonstrates superior performance in generating concise and coherent summaries when compared to existing methods. The model's rogue score on CNN/Daily Mail dataset was 44.16/21.56/41.30 on  Rogue-1, Rogue-2, and Rogue-L respectively.

\section{Methodology}
\subsection{Dataset}
The dataset that we utilized for our research is the CNN/Daily Mail dataset, a publicly accessible compilation of online news articles and accompanying summaries that originates from two of the world's most renowned news websites: CNN and Daily Mail. This dataset was compiled by a team of Google DeepMind researchers led by Karl Moritz Hermann [12].

The dataset consists of approximately 300,000  articles and their respective summaries, where the full-text news articles are in the 'article' column and their summaries in the 'highlights' column. There is another column called 'id'  containing the hexadecimal formatted SHA1 hash of the URL where the story was retrieved from. Table 1 explains each column.

Each entry in the dataset corresponds to a separate news article and is accompanied by a summary section. Typically, CNN article summaries appear as bullet points beneath the headline and before the primary text. For Daily Mail articles, executive summaries appear as "linked" sentences scattered throughout the text. Politics, international relations, business, technology, health, culture, and entertainment are just some of the topics covered by the dataset, which reflects the extensive range of topics covered by the two news organizations.

The data was compiled from publicly accessible sections of the CNN and Daily Mail websites. The data collection and preprocessing process involved scraping articles and their summaries, removing unnecessary HTML, resolving article-summary connections for Daily Mail articles, and separating the data into training, validation, and test sets. The data had 286,817 training, 13,368 validation, and 11,487 test instances.

\begin{table}[h]
\centering
\caption{Columns and Definitions}
\label{tab:defs}
\begin{tabular}{|l|p{5cm}|}
\hline
\textbf{Column Name} & \textbf{Definition} \\ \hline
id & a string containing the heximal formated SHA1 hash of the url of the article \\ \hline
article &  a string containing the body of the news article \\ \hline
highlights & a string containing the highlight of the article \\ \hline
\end{tabular}
\end{table}

\subsection{Pre-Processing of Data}
\subsubsection{Data Cleaning}
The first step in data preprocessing is to remove invalid or missing data elements from the dataset. This process involves eliminating any data points, that are missing or not required for the further processing of data.[13]
\begin{itemize}
\item{Column Dropping:}
Upon analyzing the data, we determined that the
column 'id' only contains the respective IDs of the articles, as it won't be useful in the summarization of articles, we decided to drop the column.

\item{Text Transformation:}
The text of both the 'article' and 'highlights' columns is transformed into a standardized format to facilitate seamless analysis by the model. By adopting a consistent case, the model's ability to recognize word variations is optimized, resulting in a more focused vocabulary.

\item{Expression Harmonization:}
Expressions like "don't" or "I'm" are unified into their expanded counterparts to ensure coherence throughout the text.  This process fosters clarity and coherence during data processing, promoting a cohesive output.

\item{Link Elimination:}
References to web addresses are discreetly eliminated to enhance data purity. Since web URLs do not contribute to the core context, their removal prevents unnecessary noise and supports better text summarization outcomes.

% \item{ HTML Tag Disguising:}
% Any HTML anchor tags ("\<a href\>") present in the text of 'highlights' and 'article' columns are inconspicuously transformed into spaces and subsequently discarded. HTML tags hold minimal relevance to text summarization and might disrupt summary cohesion.
\item{HTML Tag Disguising:}
Any HTML anchor tags (\textless a href\textgreater) present in the text of 'highlights' and 'article' columns are inconspicuously transformed into spaces and subsequently discarded. HTML tags hold minimal relevance to text summarization and might disrupt summary cohesion.

\item{Special Character Concealment:}
Specific characters, such as underscores, hyphens, semicolons, parentheses, and others, are subtly removed from the text. Reducing these characters from the both 'article' and 'highlight' columns minimizes data distractions and optimizes text representation.

\item{Single Quote Masking:}
Single quotes are tactfully eliminated from the text. While they serve as quotation indicators, they might not significantly influence summarization tasks. Their discreet removal from both columns fosters a sleeker text presentation.
\end{itemize}
\subsubsection{Tokenization}
Tokenization is a crucial stage in Natural Language Processing (NLP), as it facilitates the breakdown of unprocessed text into tokens. These identifiers are necessary for the processing and comprehension of human language by machines [14]. We used two Tokeinzers in our research for different methods.
\begin{itemize}
\item {Keras Tokenizer:}
It is an integral component of the Keras library,whose role is to deconstruct the text into individual words and generate a corpus-wide vocabulary. This procedure enables the NLP model to comprehend the subtleties and nuances of language at a fine-grained level [15].
The Keras Tokenizer also converts the text into sequences of integers and thus also performs the role of a vectorizer too. The tokenizer converts the text into sequences of integers, with each integer corresponding to the index of the respective word in the dictionary. We utilized the Keras Tokenizer in Abstractive Models for both 'article' and 'highlights' column as Deep Learning models and only accept integer values .

\item{ Sentence Segmentation Tokenizer:}
It is a part of the NLTK library, which is designed for segmentation at the sentence level,This tokenizer uses language-specific rules and punctuation marks to identify sentence boundaries, producing a coherent set of sentences for further analysis [16].

\end{itemize}
 \subsubsection{Vectorization}
Vectorization  involves converting textual data, such as sentences or documents, into numerical representations, known as vectors. By transforming text into numerical format, vectorization enables machine learning models to process and analyze language, as most algorithms operate on numerical data [17] . We used TF-IDF vectorizer in our extractive models and for the DeepLearning models, the vectorization was done by the Keras Tokeinzer.
\begin{itemize}
\item{TF-IDF Vectorizer:} Term frequency-inverse document frequency (TF-IDF)  vectorization [18] finds the TF (Term Frequency) for each word in a document by dividing the number of times the word appears in the document by the total number of words in the document. Then, it calculates the IDF (Inverse Document Frequency) for each word by taking the log of the total number of documents and dividing it by the number of documents that contain the word. The TF-IDF score for each word in a text is made by multiplying its TF value by its IDF value, resulting TF-IDF vectors, that shows the importance of each word in the document compared to how important it is in the whole corpus.
\end{itemize}
\subsubsection{Embedding}
Embeddings are referred to as the numerical representations of words and phrases in a lower-dimensional space. These representations record semantic and contextual information about the original data and are learned using different methods, including neural networks used for abstractive summarization. [19].
We used GloVe embeddings [20] in our study , which were Developed by Pennington et al. in 2014, GloVe is an unsupervised learning algorithm that focuses on capturing the relationships between words in a corpus.
GloVe works by analyzing word co-occurrences across the entire corpus, it uncovers semantic and contextual associations between words 
In our study, we used the 'glove.6B.100d.txt' file which contains pre-trained GloVe word 
embeddings of dimension 100.

\subsection{Automatic Text Summarization Models }
In our study, we utilized two main types of text summarization methods: extractive summarization and abstractive summarization [2][3]. Extractive summarization is the summarization where we take the most important sentences or phrases straight from the source text. On the other hand, abstractive summarization involves making a summary by understanding the context of the text and using natural language generation methods to make new sentences that convey the key information [3].

\begin{itemize}
\item \textbf{Extractive Summarization }  
  
 \subsubsection{ TF-IDF Extractive Model(Baseline):}
  The algorithm first uses the sentence tokenizer from the NLTK (Natural Language Toolkit) library to segment the raw text containing articles into individual sentences, resulting in a list of sentences. Subsequently, it calculates the TF-IDF (Term Frequency-Inverse Document Frequency) scores for each sentence, measuring word importance within the sentence relative to its significance across the entire collection. The algorithm then computes the importance of each sentence by aggregating its TF-IDF scores, generating a list of sentence importance scores. Based on these importance scores, the sentences are ranked from most to least important, allowing the identification of the key information from the input text in the top-ranked phrases. In the end, the selected sentences are combined to form the extractive summary, capturing the essential content of the original text.

 \subsubsection{Graph Based Extractive Summarizer:}
  The graph-based model [21] for extractive text summarization utilizes the cosine relationship to measure sentence similarity. Initially, the text is tokenized into sentences, and a TF-IDF vectorizer is applied, transforming sentences into numerical representations within a TF-IDF matrix. This matrix quantifies word importance in each sentence across the entire text. Subsequently, the cosine similarity is calculated between pairs of sentences, resulting in a "similarity matrix." Ranging from -1 to 1, higher values indicate greater similarity, while lower values imply dissimilarity.
Leveraging the similarity matrix, a graph is constructed with sentences as nodes and edge weights representing sentence similarity. Google's PageRank algorithm [22] is then applied to the graph, providing each sentence with a score reflecting its importance in the context of the text. By ranking the sentences based on their PageRank scores, the most significant ones are identified. Consequently, the extractive summary is composed of the top-ranked sentences, effectively capturing the essential content of the original text. This graph-based approach, integrating cosine similarity and PageRank, demonstrates promising results in the field of extractive text summarization.
\begin{figure}[htbp]
\centering
\includegraphics[width=8cm]{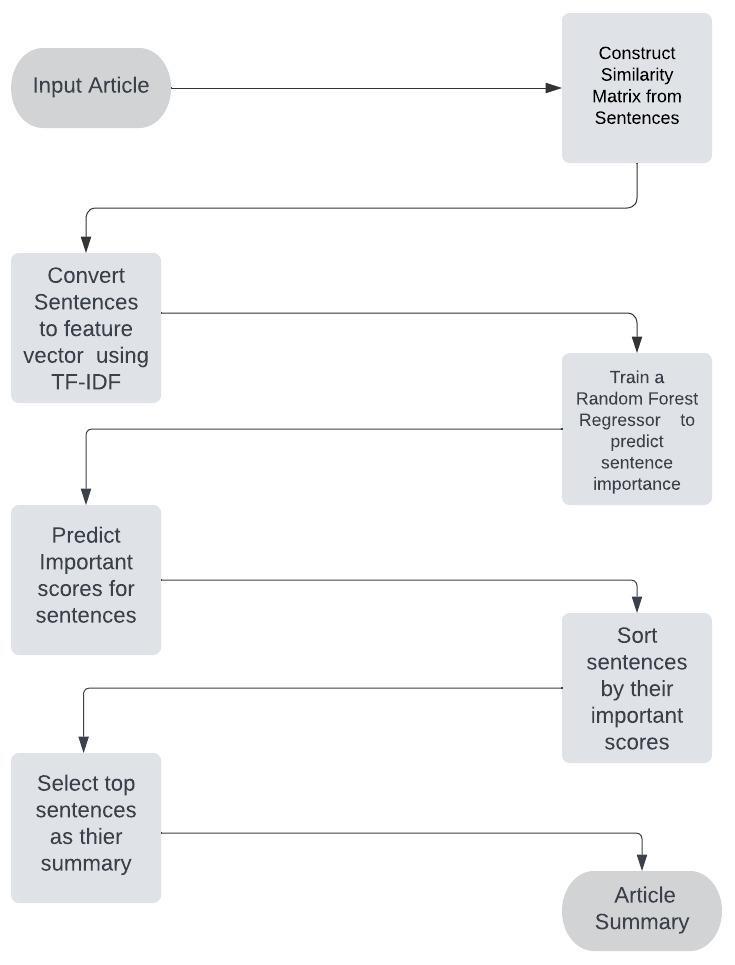}
\caption{Working of Hybrid(Extractive) Model}
\label{fig}
\end{figure}
  
\subsubsection{Hybrid}
The Hybrid model presents an algorithm that uses Google’s Page ranking Algorithm [22] and a Regressor model. It follows a similar approach as the Graph-based abstractive model, the only different thing is that here we train a regressor, a Random Forest regressor [23] to predict the sentence scores rather than taking and calculating them using the Pagerank Algorithm, we make the regressor to learn the scores from the Pagerank algorithm first and then predict the scores of all rest sentences and lastly, the sentences are put in a sequence of their respective scores, giving us the summary required.

\item\textbf{Abstractive Summarization}  

\subsubsection{Sequence-to-Sequence (Seq2Seq) model}

The Sequence-to-Sequence (Seq2Seq) model [24] is a powerful deep-learning architecture introduced in the year 2014. This sophisticated neural network is designed for tasks involving sequence mapping, which includes machine translation and text summarization.
The Seq2Seq model is composed of two essential components: an encoder and a decoder. The encoder skillfully processes the input sequence using a specialized variant called Bidirectional Long Short-Term Memory (LSTM) [25]. This LSTM captures crucial contextual information from both forward and backward directions of the input sequence. Consequently, it creates a fixed-size context vector, commonly referred to as the "thought vector," which serves as a dense representation of the input data.

The decoder utilizes the context vector from the encoder to generate the output sequence. Employing a regular LSTM, the decoder processes the context vector alongside previously generated output tokens. At each time step, the decoder effectively predicts the next token in the sequence by producing a probability distribution over the output vocabulary.

 \subsubsection{Pegasus Model}
The Pegasus model [11] is a Transformer based advanced neural network with two crucial training steps: pre-training and fine-tuning. During pre-training, it learns from a vast and diverse corpus to predict missing parts of the text. This enables Pegasus to understand language patterns and semantics effectively.
In the fine-tuning phase, Pegasus specializes in summarization using specific datasets. It learns to generate concise and coherent summaries by attending to important parts of the input. The core of Pegasus is based on the Transformer model, which handles sequential data and focuses on relevant information during generation.
\end{itemize}

\section{Experiments and Evaluation}

The purpose of this evaluation is to evaluate the performance of both extractive and abstraction models for summarization ROUGE (Recall-Oriented Understudy for Gisting Evaluation) scores [26]. ROUGE is a set of metrics that are used to 
evaluate the quality of text summarization systems. These metrics assess the similarity 
between a generated summary and one or more reference summaries. There are 3 common scores matrices in ROUGE : ROUGE-1,ROUGE-2,ROUGE-L where ROUGE-1 Measures the overlap of unigrams (single words) 
between the generated summary and the reference summary. It calculates precision, recall, 
and F1 scores based on matching individual words whereas ROUGE-2 does a similar calculation as ROUGE-1 for bigrams(pairs of consecutive 
words) and ROUGE-L computes the longest common subsequence (LCS) between the generated and 
reference summaries. We used the first avg ROUGE scores of 100 predicted summaries of the articles of the validation dataset to get the rouge scores of the different models.

\subsubsection{\textbf{Extractive Models}}
The ROUGE scores for the TF-IDF Extractive (Baseline) model are as follows: ROUGE-1 of 0.2521, ROUGE-2 of 0.0694, and ROUGE-L  of 0.2286. 

\begin{figure}[htbp]
\includegraphics[width=8cm]{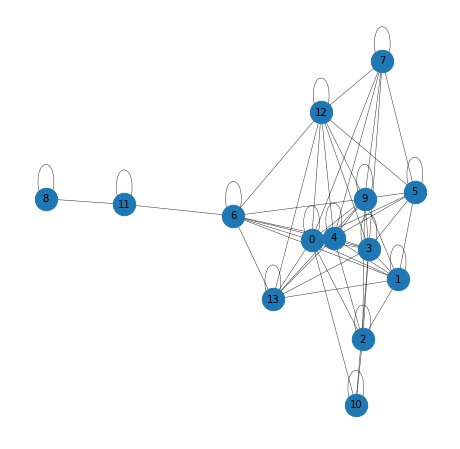}
\caption{Graph of the news article made by Graph-based Model}
\label{fig}
\end{figure}
\begin{figure}[htbp]
\includegraphics[width=8cm]{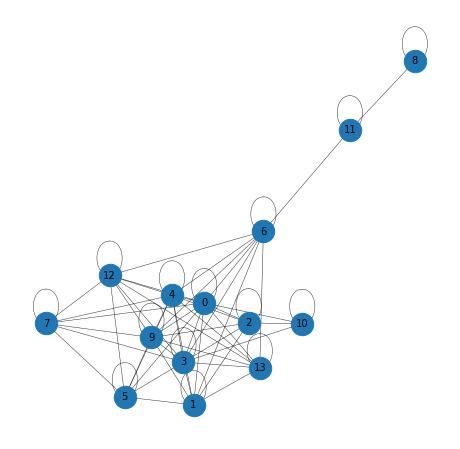}
\caption{Graph of the same article as in fig 2 made by Hybrid (Extractive) Model.}
\label{fig}
\end{figure}
In the graph-based extractive model, we take an approach where we first find the similarity score between the sentences of the news article using cosine similarity. A graph is constructed in which each node of the graph represents a sentence and the weight of the edge between two nodes is the similarity between the different sentences of the article. A sentence is considered important if it is connected to multiple nodes of the graph and then the sentences are ranked on the basis of their Google page rank scores, the top-rated sentences are then filtered and joined together to form the summary. Figure 2 shows an example of the graph model of a news article (Appendix A). The model yields ROUGE scores(ROUGE-1,ROUGE-2, ROUGE-3) of 0.332, 0.1217 , and .3044 respectively .

The Hybrid Extractive Model is similar to the graph-based model but after applying the PageRank algorithm, we use a machine learning model, a Random Forest Regressor, to predict sentence importance where the features used by the regressor are the TF-IDF vectors of the sentences of the news article. The target variable that we are trying to predict is the PageRank score and we use the predicted scores to find the importance of the sentences in the news article and thus then join the most scored ones to give a summary. The ROUGE-1, ROUGE-2, and ROUGE-L are 0.28, 0.09, and 0.262 respectively. Figure 3 shows the graph representation of the same news article(Appendix A) used in Figure 2 using the Hybrid Model.
respectively.

\subsubsection{\textbf{Abstractive Models:}}
In our first Abstractive Model, the Sequence to Sequence Model is defined with two inputs (the encoder input and the decoder input) and one output (the decoder output). The preprocessed articles in the form of integer sequences are fed into the input layer, which is then converted to dense vectors in the embedding layer, the embedded data is then sent to a Bidirectional LSTM layer which processes the input and output sequences.
The decoder layer also starts with an input sequence where these sequences again are converted to dense vectors using another embedding layer, then another LSTM layer processes the input with the context vectors from the encoder.
The output of the encoder is then fed into a TimeDistributed dense layer, which applies a softmax activation function to predict the next word in the output sequence. Figure 4 shows the blueprint of our model where 'None' mentioned in the output layers means integer type input.
\begin{figure}[htbp]
\includegraphics[width=8cm]{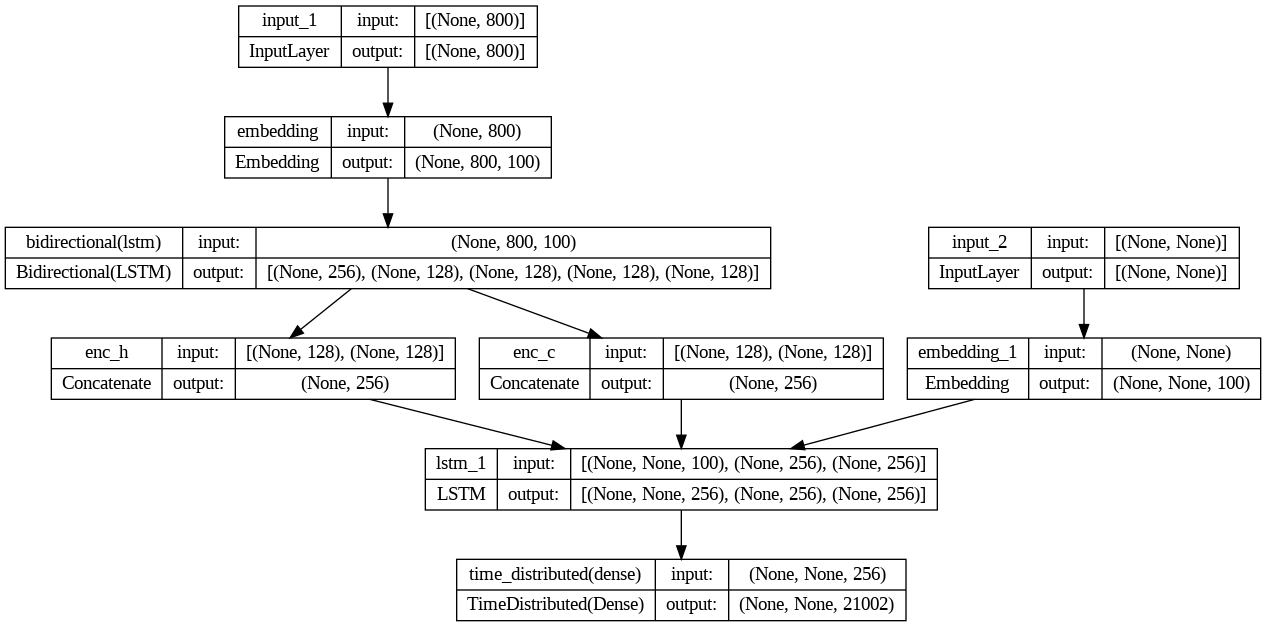}
\caption{Blueprint of the Seq2Seq Model}
\label{fig}
\end{figure}

In the training phase, we used the spars categorical cross-entropy loss function and trained our model on 10 epochs representing the 
number of times the model iterates over the entire training dataset, with a batch size of 128.
We trained our model using 55000 instances of the training dataset as the model was complex and required an additional amount of computing power to process. The ROUGE scores were as follows ROUGE-1: 0.07568, ROUGE-2: 0.00656 and ROUGE-L: 0.07568.

In our Second Abstractive Model, we used a pre-trained model called Pegasus and fine-tuned the model on our CNN/Daily News dataset, going through the data for 20 epochs with a batch size of 128. The model achieved the following ROUGE scores: ROUGE-1 of 0.3186, ROUGE-2 of 0.1269, and ROUGE-L of 0.2939.
\begin{table}[htbp]
\caption{ROUGE scores of models}
\centering
\begin{tabular}{|c|c|c|c|}
\hline
\textbf{Model} & \textbf{ROUGE-1} & \textbf{ROUGE-2} & \textbf{ROUGE-L} \\ \hline
Baseline (Extractive) & 0.2521 & 0.0694 & 0.2286 \\ \hline
Graph-Based (Extractive) & 0.332 & 0.1217 & 0.3044 \\ \hline
Hybrid (Extractive)  & 0.28 & 0.09 & 0.262 \\ \hline
Seq2Seq (Abstractive) & 0.07568 & 0.00656 & 0.07568 \\ \hline
Pegasus (Abstractive) & 0.3186 & 0.126 & 0.290 \\ \hline

\end{tabular}
\end{table}

\section{Analysis}
Extractive summarization directly selects important sentences or phrases from the original text. In the evaluation of various algorithms, the baseline extractive model that used TD-IDF scoring achieved ROUGE scores: ROUGE-1 at 0.2521, ROUGE-2 at 0.0694, and ROUGE-L at 0.2286. However, the graph-based extractive method showed better performance, outperforming the baseline with ROUGE-1 at 0.332, ROUGE-2 at 0.1217, and ROUGE-L at 0.3044. Additionally, a hybrid algorithm, which used a regressor to predict the sentence score got promising results with ROUGE-1 at 0.28, ROUGE-2 at 0.09, and ROUGE-L at 0.262 scores but the best performing was the Graph-Based Model.

On the other hand, abstractive summarization model summaries use natural language and generally demand more computational resources than extractive methods. The Seq2Seq  model got the lowest ROUGE scores among all algorithms: ROUGE-1 at 0.07568, ROUGE-2 at 0.00656, and ROUGE-L at 0.07568. Its model lacked performance due to inadequate computation power. On the other hand, the Pegasus abstractive model achieved relatively favorable results with ROUGE-1 at 0.313505, ROUGE-2 at 0.126, and ROUGE-L at 0.290 scores. 

It is important to note that abstractive models, like Seq2Seq and Pegasus are more resource-intensive compared to extractive ones. Abstractive models often require large amounts of computation power and memory to process and generate summaries effectively, explaining the relatively lower ROUGE scores of the Seq2Seq model.

Similarly, the Pegasus model can achieve even higher ROUGE scores if trained with more complex approaches, but the limitation of computational power acts as a constraint. 

\section{Application}
We developed a web application to test the Models with the Live News Articles provided by an API called - NewsAPI [27],
The front end of the application is created using HTML (Hypertext Markup Language) which allowed the creation of the basic structure and content of the application, and CSS (Cascading Style Sheets) to enhance the design and layout of the webpage[28][29] (Appendix B).

The backend of the application is based on Flask [30], a lightweight and flexible web framework written in Python programming language. Flask provides a library and a collection of codes that can be used to build websites.

We used the best 3 Models from the study and tested them on the Live news Articles that came through the NewsAPI and checked which method takes the longest time to process the news in the Real-Time Summarization of news Articles, It was found that Extractive Models take far less time than the Pegasus Model for execution in real-time as shown in Figure 5.
\begin{figure}[htbp]
\includegraphics[width=8cm]{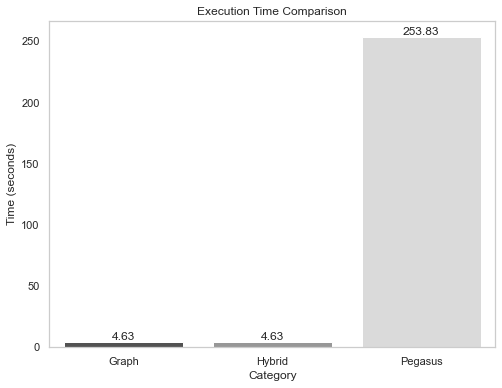}
\caption{Exectuion Time of best models on the Web Application}
\label{fig}
\end{figure}

\section{Conclusion}

In this study, we looked at both extractive and abstractive methods for summarising text using the CNN Daily Mail dataset. The pre-trained PEGASUS model had the highest ROUGE score of all the models that were tried, showing that it was the best at summarising. This is because of its sophisticated transformer-based architecture and thorough pre-training on a wide range of text corpus.

However, the practical application of these models in real-time scenarios to summarize live news poses unique challenges. When it comes to speed and efficiency, the extractive models are much better than their abstractive peers. So, they are a natural choice when you need to quickly summarise the text.

It is important to note that the full potential of abstractive models could not be used because they require a lot of computing power to train.

In the future, work in this area can look at many different things. One of the most important things could be to improve how abstract models are implemented so that they can be used in real-time apps without hurting the quality of the summaries they produce. Also, looking into ways to train these models using distributed computing or quantum computing in an efficient way can be a good idea. Also, more studies can be done to come up with new architectures or training methods that balance performance and efficiency of computation. 

In conclusion, study into text summarization has come a long way, but there is still a lot to learn and improve, especially for real-time applications. This field has an exciting future because hardware capabilities and machine learning methods are always getting better.

\vspace{12pt}
\appendix

\subsection{Example Summary of a news Article}

Original Article:
 A drunk teenage boy had to be rescued by security after jumping into a lions' enclosure at a zoo in western India. Rahul Kumar, 17, clambered over the enclosure fence at theÂ Kamla Nehru Zoological Park in Ahmedabad, and began running towards the animals, shouting he would 'kill them'. Mr Kumar explained afterwards that he was drunk and 'thought I'd stand a good chance' against the predators. Next level drunk: Intoxicated Rahul Kumar, 17, climbed into the lions' enclosure at a zoo in Ahmedabad and began running towards the animals shouting 'Today I kill a lion!' Mr Kumar had been sitting near the enclosure when he suddenly made a dash for the lions, surprising zoo security. The intoxicated teenager ran towards the lions, shouting: 'Today I kill a lion or a lion kills me!' A zoo spokesman said: 'Guards had earlier spotted him close to the enclosure but had no idea he was planning to enter it. 'Fortunately, there are eight moats to cross before getting to where the lions usually are and he fell into the second one, allowing guards to catch up with him and take him out. 'We then handed him over to the police.' Brave fool: Fortunately, Mr Kumar  fell into a moat as he ran towards the lions and could be rescued by zoo security staff before reaching the animals (stock image) Kumar later explained: 'I don't really know why I did it. 'I was drunk and thought I'd stand a good chance.' A police spokesman said: 'He has been cautioned and will be sent for psychiatric evaluation. 'Fortunately for him, the lions were asleep and the zoo guards acted quickly enough to prevent a tragedy similar to that in Delhi.' Last year a 20-year-old man was mauled to death by a tiger in the Indian capital after climbing into its enclosure at the city zoo.
\begin{itemize}

\item \textbf{Generated Summary using Graph Model :}

 Next level drunk: Intoxicated Rahul Kumar, 17, climbed into the lions' enclosure at a zoo in Ahmedabad and began running towards the animals shouting 'Today I kill a lion!' Brave fool: Fortunately, Mr Kumar  fell into a moat as he ran towards the lions and could be rescued by zoo security staff before reaching the animals (stock image) Kumar later explained: 'I don't really know why I did it. Mr Kumar explained afterward that he was drunk and 'thought I'd stand a good chance' against the predators

 \item \textbf{Generated Summary using Hybrid Model :}

 Next level drunk: Intoxicated Rahul Kumar, 17, climbed into the lions' enclosure at a zoo in Ahmedabad and began running towards the animals shouting 'Today I kill a lion!' Brave fool: Fortunately, Mr Kumar  fell into a moat as he ran towards the lions and could be rescued by zoo security staff before reaching the animals (stock image) Kumar later explained: 'I don't really know why I did it.
 
 \item \textbf{Generated Summary using Pegasus Model :}

 Next level drunk: Intoxicated Rahul Kumar, 17, climbed into the lions' enclosure at a zoo in Ahmedabad and began running towards the animals shouting 'Today I kill a lion!' Mr Kumar had been sitting near the enclosure when he suddenly made a dash for the lions, surprising zoo security.
 \subsection{Web Application}
 \begin{figure}[htbp]
\centering
\includegraphics[width=8cm]{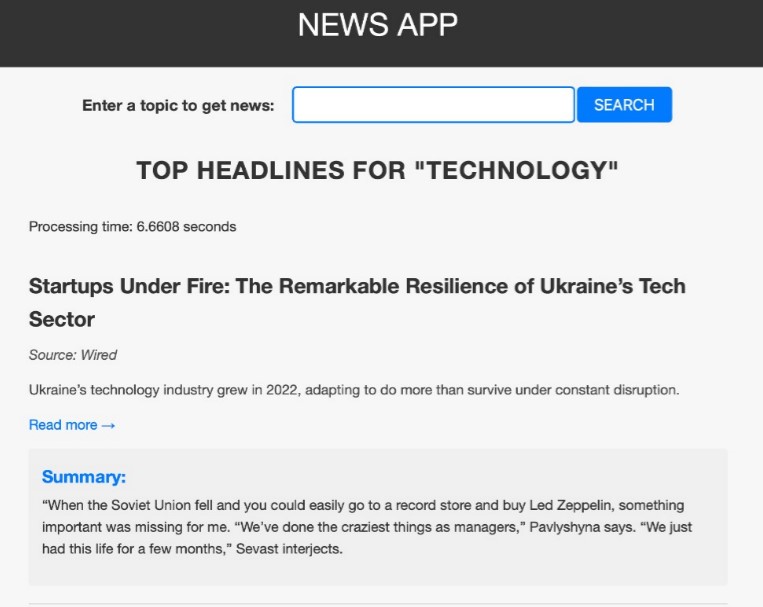}
\caption{Screenshot of the Application.}
\label{fig}
\end{figure}
 \end{itemize}
\end{document}